\documentclass[
twocolumn,
]{ceurart}

\usepackage{listings}
\usepackage{flushend}
\usepackage{booktabs}
\usepackage{adjustbox}
\usepackage{enumitem}
\usepackage{multirow}
\usepackage{verbatim}
\usepackage{xcolor}
\usepackage[most]{tcolorbox}
\usepackage{newunicodechar}
\newunicodechar{↔}{\ensuremath{\leftrightarrow}}
\newcommand{\algoname}[1]{{\textsc{#1}}\xspace}
\newcommand{\ourmodel}{\algoname{DIETA}}
\newcommand{\ourmodelsynth}{\algoname{DIETA\textsubscript{+BT}}}

\begin{document}

\copyrightyear{2025}
\copyrightclause{Copyright for this paper by its authors.
  Use permitted under Creative Commons License Attribution 4.0
  International (CC BY 4.0).}

\conference{CLiC-it 2025: Eleventh Italian Conference on Computational Linguistics, September 24 — 26, 2025, Cagliari, Italy}

\title{DIETA: A Decoder-only transformer-based model for Italian–English machine TrAnslation}


\address[1]{Department of Informatics, Systems and Communication - DISCo, University of Milano-Bicocca, Italy}
\address[2]{DAUIN Dipartimento di Automatica e Informatica, Politecnico di Torino, Italy}
\address[3]{NVIDIA AI Technology Center, Italy}
\address[4]{Università degli Studi di Pavia, Italy}

\author[1]{Pranav Kasela}[%
orcid=0000-0003-0972-2424,
email=pranav.kasela@unimib.it,
url=https://pkasela.github.io/,
]

\author[1,2]{Marco Braga}[%
orcid=0009-0004-7619-8399,
email=m.braga@campus.unimib.it,
]

\author[4]{Alessandro Ghiotto}[%
email=alessandro.ghiotto01@universitadipavia.it,
]


\author[3]{Andrea Pilzer}[%
orcid=0000-0001-6868-7943,
email=apilzer@nvidia.com,
]

\author[1]{Marco Viviani}[%
orcid=0000-0002-2274-9050,
email=marco.viviani@unimib.it,
]

\author[1]{Alessandro Raganato}[%
orcid=0000-0002-7018-7515,
email=alessandro.raganato@unimib.it,
]


\begin{abstract}
In this paper, we present \textbf{DIETA}, a small, decoder-only Transformer model with 0.5 billion parameters, specifically designed and trained for Italian–English machine translation. We collect and curate a large parallel corpus consisting of approximately 207 million Italian–English sentence pairs across diverse domains, including parliamentary proceedings, legal texts, web-crawled content, subtitles, news, literature and 352 million back-translated data using pretrained models. Additionally, we create and release a new small-scale evaluation set, consisting of 450 sentences, based on 2025 WikiNews articles, enabling assessment of translation quality on contemporary text. Comprehensive evaluations show that DIETA achieves competitive performance on multiple Italian–English benchmarks, consistently ranking in the second quartile of a 32-system leaderboard and outperforming most other sub-3B models on four out of five test suites. The training script, trained models, curated corpus, and newly introduced evaluation set are made publicly available, facilitating further research and development in specialized Italian–English machine translation: \href{https://github.com/pkasela/DIETA-Machine-Translation}{https://github.com/pkasela/DIETA-Machine-Translation}.

\end{abstract}


\maketitle

\section{Introduction}

Transformer-based Large Language Models (LLMs) have significantly advanced Natural Language Processing (NLP) tasks, such as Text Classification \cite{braga-etal-2024-adakron}, community question answering \cite{braga2025synthetic, kasela_sepqa_2024}, health applications including clinical trial retrieval and automated psychiatric assessment  \cite{peikos24leveraging,raganato2024leveraging}, and Machine Translation (MT) \cite{vaswani2017attention,survey_llms,hendy2023good}. Despite these versatile applications, the problem of high‑quality neural machine translation (MT), especially for language pairs like Italian–English, remains an open challenge. General-purpose multilingual systems often prioritize broad coverage over specialized translation quality, leaving significant room for improvement in targeted language pairs.

To address these limitations, we introduce \textbf{DIETA}, a small decoder-only Transformer model with 0.5 billion parameters, specifically tailored for high-quality bidirectional Italian–English translation. Furthermore, we compiled an extensive parallel corpus consisting of approximately 207 million high-quality bilingual sentence pairs from publicly accessible resources, including parliamentary records (Europarl \cite{koehn2005europarl}, DGT-TM \cite{TIEDEMANN12.463}), legal texts, web-crawled content (ParaCrawl \cite{espla2019paracrawl}), subtitles (OpenSubtitles \cite{lison2018opensubtitles2018,lison-tiedemann-2016-opensubtitles2016}), and encyclopedic and literary sources (WikiMatrix \cite{schwenk2021wikimatrix}, Books). In order to recognize the importance of linguistic diversity and temporal relevance, we augmented this dataset by constructing an additional synthetic corpus of 352 million sentence pairs via back-translation, specifically targeting news-related content.

To evaluate \ourmodel's performance on recent domains, we created and released a small-scale evaluation set, \textbf{WikiNews-25}, based on 2025 WikiNews articles. This dataset consists of post-edited translations, carefully selected to include only those segments that initially contained translation errors requiring human correction. 
Our experimental comparisons include multilingual models (e.g., NLLB-200 \cite{costa2022no}) and Italian-English models (e.g., OPUS-MT \cite{tiedemann2024democratizing,tiedemann-thottingal-2020-opus}, Minerva \cite{orlando2024minerva}, LLaMAntino \cite{basile2023llamantino}) across five established benchmarks. We compared several model variants, trained with and without synthetic back-translated data. Results show that \ourmodel consistently ranks in the second quartile among 32 evaluated systems, outperforming all comparable models below 3 billion parameters on four out of five test suites, while requiring less GPU memory than larger multilingual baselines.

In summary, our main contributions include:
(i) training and releasing a specialized, small decoder-only Transformer model optimized for high-quality Italian–English translation;
(ii) creating and publicly releasing a large-scale, carefully curated parallel corpus from diverse sources, and generating a synthetic corpus through back-translation;
(iii) introducing the new WikiNews-25 evaluation set to facilitate benchmarking on recent, human-corrected content;
(iv) conducting thorough evaluations using multiple MT metrics.

\section{Related Works}
\label{Sec:RW}

Publicly available bilingual corpora play a central role in the development and evaluation of Machine Translation (MT) systems. Among these, OPUS \cite{tiedemann-2016-opus,tiedemann2024democratizing} is a well-known source of multilingual datasets that have been widely used in both statistical and neural MT research. Large-scale web-crawled corpora such as ParaCrawl \cite{espla2019paracrawl} and NLLB \cite{fan2021beyond} are particularly noteworthy for their coverage and scale, making them important resources for training state-of-the-art multilingual MT models.

Recent Transformer models such as \mbox{mBART-50} \cite{tang2020multilingual}, \mbox{NLLB-200} \cite{fan2021beyond}, MADLAD-400 \cite{kudugunta2023madlad400}, Tower \cite{alves2024tower}, and Gemma-2 \cite{team2024gemma} have showed that expanding language coverage and model capacity can significantly enhance many-to-many translation quality. However, the computational demands of these massive models, and the inherent competition for representational capacity across hundreds of languages, often leave room for improvement on specific language pairs such as English–Italian. For many language directions, the open OPUS-MT family \cite{tiedemann-thottingal-2020-opus,tiedemann2024democratizing} remains a widely used baseline, yet its more compact architectures lag behind the newest LLM-based systems in fluency and versatility.

General-purpose models like the GPT and LLaMA series, when prompted or instruction-tuned, achieve impressive zero-shot MT results. Specialised variants, like GemmaX2-28 \cite{cui2025multilingualmachinetranslationopen}, further narrow the gap with commercial MT engines. Meanwhile, to strengthen the representation of Italian within multilingual LLMs, several initiatives have introduced Italian-focused systems. Models such as LLaMAntino \cite{basile2023llamantino}, Minerva \cite{orlando-etal-2024-minerva}, Cerbero \cite{galatolo2023cerbero},  ModelloItalia \cite{Modello_Italia}, and DanteLLM \cite{bacciu-etal-2024-dantellm} leverage hundreds of billions of Italian tokens and human feedback to yield substantial improvements in Italian generation and understanding. Nonetheless, these models are designed as general-purpose language models and are not optimised specifically for the MT task.

In this work, we introduce a compact, 0.5B-parameter decoder-only model, trained from scratch on a total of \emph{768 million} parallel and synthetic sentence pairs, delivering a purpose-built, open solution for English$\leftrightarrow$Italian machine translation.


\section{Data Collection and Preparation}
\label{Sec:Data}

This section outlines the creation of a large Italian–English sentence pair corpus and a synthetic dataset derived from Web News and crawled data.

\subsection{Parallel Training Corpus}
To build a decoder‑only model for bidirectional \textit{English ↔ Italian} translation, we make use of every public bitext for the pair available in OPUS \cite{tiedemann-2016-opus}. Sources span Web crawls \cite{schwenk-etal-2021-ccmatrix,fan2021beyond,espla2019paracrawl}, Wikipedia \cite{TIEDEMANN12.463,wolk2014building,schwenk2021wikimatrix}, parliamentary/legal proceedings \cite{koehn2005europarl,steinberger2006jrc}, and film/TV subtitles \cite{lison2018opensubtitles2018}.  
Because the NLLB corpus \cite{fan2021beyond} contains CCMatrix, we keep only the NLLB portion to prevent duplication.

\paragraph{Cleaning and quality control.}
We remove exact duplicates using \textsc{OpusTools} and \textsc{OpusFilter} \cite{aulamo2020opustools,aulamo-etal-2020-opusfilter}, then pass each remaining sentence pair to the Phi‑4 LLM \cite{abdin2024phi} with the binary prompt shown in Figure~\ref{fig:prompt_filter}. Pairs that receive \texttt{no} are discarded.

\begin{figure}[ht]
\centering
\begin{tcolorbox}[colback=gray!10,colframe=black!70,title=Filtering prompt]
\small
Given the English and Italian sentences below, are they translations of each other?  
Answer with yes or no only.
\end{tcolorbox}
\caption{Prompt issued to \textsc{Phi‑4} during quality filtering.}
\label{fig:prompt_filter}
\end{figure}

After cleaning, the corpus contains $\mathbf{207~864~437}$ high-quality sentence pairs. For bidirectional training, each pair is duplicated with explicit direction tags, resulting in a total of $\mathbf{415~728~874}$ source–target examples, as illustrated in Figure~\ref{fig:template}.

\begin{figure}[ht]
\centering
\begin{tcolorbox}[colback=gray!10,colframe=black!70,title=Sample formatting]
\footnotesize
\texttt{ENG:} \textit{English sentence} \texttt{IT:} \textit{Italian translation} \\[3pt]
\texttt{IT:} \textit{Italian sentence} \texttt{ENG:} \textit{English translation}
\end{tcolorbox}
\caption{Sample formatting with explicit language tags used for training the \ourmodel models.}
\label{fig:template}
\end{figure}


\subsection{Synthetic Data via Back‑Translation}
To expand the parallel training corpus, we generated additional sentence pairs by back‑translation \cite{sennrich2016improving}. As monolingual sources we used the \textsc{NewsCrawl}\footnote{\url{https://data.statmt.org/news-crawl/}} corpora \cite{kocmi-etal-2024-findings} and the web‑scale \textsc{FineWeb} collection \cite{penedo2024the,penedo2025fineweb2pipelinescale}.

\paragraph{NewsCrawl.} We translated Italian articles from 2008–2018 and English articles from 2023 with the \texttt{OPUS-MT-TC-BIG} model \cite{tiedemann-thottingal-2020-opus,tiedemann-2020-tatoeba,tiedemann2024democratizing}. The remaining segments (Italian 2019–2024 and English 2024) were translated with \texttt{NLLB-200-3.3B} \cite{nllb2022}. In total, this yielded \textbf{144,189,087} synthetic sentence pairs, comprising 67.8~M Italian and 76.3~M English sentences.

\paragraph{FineWeb.} From the multilingual \textsc{FineWeb2} we translated 108.5~M Italian sentences, and from the English \textsc{FineWeb} crawl we translated 100~M English sentences resulting in a total of \textbf{208,516,318} sentences, using the multilingual \texttt{GemmaX2‑28‑9B‑v0.1} model \cite{cui2025multilingualmachinetranslationopen}.

All translations were generated with the \texttt{CTranslate2}\footnote{\url{https://github.com/OpenNMT/CTranslate2}} toolkit in greedy decoding mode for efficient inference with large Transformer models.

\subsection{Training corpus summary} 
Duplicating the OPUS parallel pairs to cover both translation directions (i.e., from English to Italian and vice versa) yields \textbf{415,728,874} direction‑specific examples. When combined with the \textbf{144,195,695} NewsCrawl and \textbf{208,516,318} FineWeb synthetic pairs, the total training set comprises \textbf{768,440,887} source–target examples. We shuffle the corpus once before mini‑batch construction.

%

\subsection{Evaluation Sets}
In addition to standard benchmarks, we release \textbf{WikiNews‑25}, a 450-segment test set based on 2025 WikiNews sentences. Machine translations generated by Google Translate were post-edited using English as the source language, retaining only those sentences that required substantive corrections.

\section{Methodology}
\label{sec:methodology}

This section describes the tokenizer, the model architecture, and the training strategy adopted to develop our proposed models.

\paragraph{Tokenizer.}
We use the 51,200-entry SentencePiece vocabulary from the \textit{Minerva} family of models \cite{orlando-etal-2024-minerva}.\footnote{\texttt{sapienzanlp/Minerva-7B-instruct-v1.0}} 
Unlike general-purpose multilingual tokenizers, Minerva's vocabulary was specifically trained on a balanced corpus of high-quality Italian and English texts, resulting in optimized sub-word segments aligned closely to the morphological and orthographic structures of both languages. 
This choice ensures that our models effectively capture nuances specific to the Italian–English language pair.

\paragraph{Model Architecture.}
\ourmodel\ is a decoder-only Transformer composed of six identical layers, each adopting a post-norm configuration. Every layer features a hidden dimension of 2048 and 32 attention heads. The feed-forward sub-layer uses a squared-ReLU activation and expands the hidden representation by a factor of four before projecting it back to the residual stream. Token positions are encoded using rotary embeddings \cite{su2024roformer}. The architecture further incorporates residual attention accumulation \cite{he2021realformer} and query-key normalization \cite{henry2020query,dehghani2023scaling}.

\paragraph{Training Schedule.}
Our models are implemented using the \textsc{x-Transformers} framework.\footnote{\url{https://github.com/lucidrains/x-transformers}} Training is performed for a single epoch over the dataset described in Section~\ref{Sec:Data}, utilizing the Lion optimizer \cite{chen2023symbolic} with a learning rate of $2\times10^{-4}$ and a linear decay schedule preceded by a warm-up phase covering the first 10\% of training steps. We release five variants of our trained model checkpoints: 
\begin{itemize}[leftmargin=*, noitemsep, topsep=2pt]
    \item \ourmodel: trained from scratch on the high-quality filtered parallel corpus (415.7M sentence pairs).
    \item \ourmodelsynth: trained on the parallel corpus plus NewsCrawl back‑translations (total 559,924,569 pairs).
    \item \algoname{DIETA\textsubscript{+cont}}: continues \ourmodel for a second epoch on the same 559,924,569‑pair mixture.
    \item \algoname{DIETA\textsubscript{+nosynth}}: continues \ourmodel for a second epoch on the original parallel data only.
    \item \algoname{DIETA\textsubscript{+allsynth}}: continues \algoname{DIETA\textsubscript{+cont}} for a third epoch on the full corpus (parallel + NewsCrawl + FineWeb), totalling 768,440,887 pairs.
\end{itemize}

\section{Experimental Setup}
\label{sec:exp_setup}




We evaluate a broad range of translation systems, providing for each the parameter count, model architecture, and main language coverage:

\begin{itemize}[leftmargin=*]
    \item \textbf{EuroLLM-1.7B} (\textit{utter-project/EuroLLM-1.7B-Instruct};
          $1.7$ B, LLaMA-style dense Transformer)\,—
          trained on \(\sim\!4\) T multilingual tokens and
          instruction-tuned on \emph{EuroBlocks}; covers 35 EU + major languages;
    \item \textbf{EuroLLM-9B} (\textit{utter-project/EuroLLM-9B-Instruct};
          $9.15$ B)\,—same recipe as above at larger scale;
    \item \textbf{LLaMAntino-8B} (\textit{swap-uniba/LLaMAntino-3-ANITA-8B-Inst-DPO-ITA};
          $8$ B, Meta-Llama-3 backbone)\,—
          EN ↔ IT instruction + DPO tuned; 
    \item \textbf{Maestrale v0.4} (\textit{mii-llm/maestrale-chat-v0.4-beta};
          $7.2$ B, Mistral-7B continued-pretrain + SFT + DPO on 1.7 M Italian
          instructions); 
    \item \textbf{mBART-50} (\textit{facebook/mbart-large-50-many-to-many-mmt};
          $0.61$ B seq-to-seq Transformer)\,—50-language many-to-many MT;
    \item \textbf{Minerva-7B} (\textit{sapienzanlp/Minerva-7B-instruct-v1.0};
          $7$ B, Mistral-like)\,—pre-trained on 2.5 T tokens
          (50 \% IT, 50 \% EN) + safety tuning;
    \item \textbf{PhiMaestra-3} (\textit{LeonardPuettmann/PhiMaestra-3-Translation};
          $3.8$ B, Phi-3 mini)\,—fine-tuned on 0.5 M \textsc{Tatoeba}
          EN↔IT pairs; 
    \item \textbf{Cerbero-7B} (\textit{galatolo/cerbero-7b};
          $7$ B, Mistral-7B base)\,—Italian-centric LLM trained on synthetic
          Cerbero corpus;
    \item \textbf{NLLB-200 (600 M / 1.3 B / 3.3 B)} (\textit{facebook/nllb-200-*})\,
          Transformer family covering 200 languages; 
    \item \textbf{opus-mt (small)} EN$\rightarrow$IT / IT$\rightarrow$EN
          (\textit{Helsinki-NLP/opus-mt-*}; $\sim\!270$ M, Marian-Transformer);
    \item \textbf{opus-mt-big} EN$\rightarrow$IT / IT$\rightarrow$EN
          (\textit{Helsinki-NLP/opus-mt-tc-big-*};
          $\sim\!560$ M Transformer model with back-translation);
    \item \textbf{ModelloItalia-9B} (\textit{sapienzanlp/modello-italia-9b};
          $9$ B, GPT-NeoX)\,—Italian LLM by iGenius/CINECA; 
    \item \textbf{Llama-3.1-8B-ITA} (\textit{DeepMount00/Llama-3.1-8b-ITA};
          $8$ B, Meta-Llama-3.1 fine-tuned for Italian);
    \item \textbf{Tower-7B} (\textit{Unbabel/TowerInstruct-7B-v0.2};
          $6.7$ B, LLaMA-2 base)\,—10-language MT and post-editing tasks;
    \item \textbf{Gemma-2B / 9B} (\textit{ModelSpace/GemmaX2-28-\{2B,9B\}};
          $3.2$ B / $10.2$ B, Gemma-2 continued-pretrain + MT SFT for 28
          languages); 
    \item \textbf{MADLAD-3B / 7B}
          (\textit{google/madlad400-\{3b,7b\}-mt};
          $3$ B / $7.2$ B, T5)\,—400+-language MT trained on up to
          1 T tokens.
\end{itemize}

\paragraph{Automatic metrics.}

\begin{table*}[ht]
\caption{NTREX-128 Translation Results. The suffix -b5 indicates that beam search with 5 beams was used during generation.}
\label{NTREX}
\centering
\resizebox{\textwidth}{!}{%
\begin{tabular}{l *{7}{cc}}
\toprule
\multirow{2}{*}{Model} & \multicolumn{2}{c}{sacrebleu($\uparrow$)} & \multicolumn{2}{c}{chrf($\uparrow$)} & \multicolumn{2}{c}{bleurt($\uparrow$)} & \multicolumn{2}{c}{metricx($\downarrow$)} & \multicolumn{2}{c}{comet($\uparrow$)} & \multicolumn{2}{c}{qemetricx($\downarrow$)} & \multicolumn{2}{c}{cometkiwi($\uparrow$)} \\
\cmidrule(lr){2-3} \cmidrule(lr){4-5} \cmidrule(lr){6-7} \cmidrule(lr){8-9} \cmidrule(lr){10-11} \cmidrule(lr){12-13} \cmidrule(lr){14-15}
& en->it & it->en & en->it & it->en & en->it & it->en & en->it & it->en & en->it & it->en & en->it & it->en & en->it & it->en \\
\midrule
Cerbero-7B	&	29.7079	&	30.9760	&	57.3078	&	57.1923	&	0.1096	&	0.0215	&	3.5111	&	4.7368	&	0.8362	&	0.8467	&	3.2268	&	4.0855	&	0.7057	&	0.6427	\\
EuroLLM-1.7B	&	20.4871	&	26.4106	&	51.9333	&	56.7543	&	0.0146	&	0.1282	&	4.3742	&	4.4232	&	0.8061	&	0.8299	&	3.7145	&	3.7631	&	0.6428	&	0.7023	\\
EuroLLM-9B	&	27.0934	&	32.2015	&	57.2185	&	60.8868	&	0.1041	&	0.3495	&	2.5383	&	3.0920	&	0.8560	&	0.8636	&	2.2266	&	2.7051	&	0.7450	&	0.7557	\\
Gemma-2B	&	44.6901	&	46.5254	&	68.0057	&	68.5879	&	0.2778	&	0.4485	&	1.8038	&	2.6064	&	0.8902	&	0.8847	&	1.8589	&	2.5327	&	0.7861	&	0.7708	\\
Gemma-2B-b5	&	45.7915	&	47.1590	&	68.9844	&	68.6866	&	0.2932	&	0.4527	&	1.6976	&	2.5509	&	0.8934	&	0.8851	&	\textbf{1.7898}	&	2.4879	&	0.7925	&	0.7723	\\
Gemma-9B	&	\textbf{50.7462}	&	48.1639	&	71.7725	&	69.7526	&	0.3523	&	0.4703	&	1.6551	&	2.4812	&	0.8992	&	0.8874	&	1.8245	&	2.5546	&	0.7912	&	0.7693	\\
Gemma-9B-b5	&	50.4767	&	\textbf{49.2683}	&	\textbf{72.4682}	&	\textbf{70.3634}	&	\textbf{0.3596}	&	\textbf{0.4787}	&	\textbf{1.6006}	&	\textbf{2.4325}	&	\textbf{0.9010}	&	\textbf{0.8888}	&	1.7965	&	2.5363	&	\textbf{0.7933}	&	0.7695	\\
Llama-3.1-8B	&	31.1660	&	41.2522	&	58.9875	&	64.7492	&	0.1484	&	0.2846	&	2.7722	&	3.4054	&	0.8589	&	0.8720	&	2.5072	&	3.1498	&	0.7510	&	0.7194	\\
LLaMAntino-8B	&	24.7926	&	34.0440	&	53.7380	&	62.1239	&	0.0606	&	0.3300	&	3.9447	&	3.1905	&	0.8198	&	0.8589	&	3.5144	&	2.8648	&	0.6846	&	0.7529	\\
Madlad-3B	&	37.8887	&	41.7829	&	63.4694	&	66.0737	&	0.2181	&	0.4264	&	2.4718	&	2.7272	&	0.8687	&	0.8790	&	2.3323	&	2.5132	&	0.7598	&	0.7734	\\
Madlad-3B-b5	&	38.4722	&	41.5983	&	64.0904	&	66.2366	&	0.2255	&	0.4246	&	2.4614	&	2.6997	&	0.8687	&	0.8785	&	2.3322	&	2.4917	&	0.7608	&	0.7744	\\
Madlad-7B	&	38.4578	&	42.5244	&	63.7821	&	66.6484	&	0.2214	&	0.4369	&	2.3396	&	2.6337	&	0.8707	&	0.8811	&	2.2737	&	2.4991	&	0.7634	&	0.7736	\\
Madlad-7B-b5	&	38.9525	&	42.2828	&	64.3319	&	66.7757	&	0.2293	&	0.4367	&	2.3629	&	2.5962	&	0.8716	&	0.8812	&	2.2685	&	\textbf{2.4546}	&	0.7635	&	\textbf{0.7757}	\\
Maestrale-v0.4	&	26.4776	&	32.4728	&	56.4607	&	60.1570	&	0.1038	&	0.2952	&	2.6550	&	3.2991	&	0.8510	&	0.8585	&	2.3368	&	2.9898	&	0.7429	&	0.7362	\\
mBART	&	29.7014	&	34.9348	&	57.4304	&	61.1602	&	0.1415	&	0.3029	&	4.1793	&	3.9910	&	0.8268	&	0.8479	&	3.6291	&	3.3345	&	0.6878	&	0.7294	\\
mBART-b5	&	29.7014	&	34.9348	&	57.4304	&	61.1602	&	0.1415	&	0.3029	&	4.1793	&	3.9910	&	0.8268	&	0.8479	&	3.6291	&	3.3345	&	0.6878	&	0.7294	\\
Minerva-7B	&	30.2021	&	25.7506	&	58.7382	&	52.6011	&	0.1320	&	-0.2292	&	2.8985	&	7.1023	&	0.8528	&	0.7727	&	2.6651	&	6.6963	&	0.7286	&	0.5846	\\
ModelloItalia-9B	&	36.2878	&	34.6847	&	62.0944	&	61.3331	&	0.1864	&	0.2572	&	2.4967	&	3.6490	&	0.8628	&	0.8548	&	2.3314	&	3.1396	&	0.7418	&	0.7192	\\
NLLB-1.3B	&	36.0274	&	42.2195	&	62.2182	&	66.3278	&	0.1985	&	0.4197	&	2.6634	&	2.8303	&	0.8617	&	0.8754	&	2.4676	&	2.6013	&	0.7532	&	0.7680	\\
NLLB-1.3B-b5	&	36.8762	&	43.0356	&	63.1081	&	66.8704	&	0.2096	&	0.4267	&	2.4641	&	2.7521	&	0.8663	&	0.8768	&	2.2992	&	2.5401	&	0.7648	&	0.7707	\\
NLLB-3.3B	&	36.5066	&	43.7135	&	62.6141	&	67.2720	&	0.2093	&	0.4306	&	2.5183	&	2.7114	&	0.8663	&	0.8774	&	2.3542	&	2.5379	&	0.7583	&	0.7698	\\
NLLB-3.3B-b5	&	37.4447	&	44.0335	&	63.4329	&	67.6084	&	0.2212	&	0.4340	&	2.3616	&	2.6609	&	0.8695	&	0.8780	&	2.2230	&	2.4876	&	0.7660	&	0.7727	\\
NLLB-600M	&	34.2615	&	40.0278	&	60.9701	&	64.7658	&	0.1860	&	0.3855	&	3.2779	&	3.1761	&	0.8466	&	0.8655	&	2.9786	&	2.7883	&	0.7233	&	0.7583	\\
NLLB-600M-b5	&	35.0643	&	40.6537	&	61.8143	&	65.1725	&	0.1968	&	0.3949	&	2.9996	&	3.0685	&	0.8546	&	0.8679	&	2.7120	&	2.7057	&	0.7389	&	0.7632	\\
opus-mt	&	32.6806	&	36.0435	&	60.1638	&	62.7542	&	0.1692	&	0.3461	&	4.1174	&	3.4280	&	0.8220	&	0.8565	&	3.7494	&	2.9983	&	0.6762	&	0.7540	\\
opus-mt-b5	&	32.7081	&	36.0080	&	60.1931	&	62.7413	&	0.1690	&	0.3458	&	4.1173	&	3.4471	&	0.8215	&	0.8563	&	3.7607	&	3.0080	&	0.6765	&	0.7537	\\
opus-mt-big	&	36.1768	&	41.5136	&	62.2987	&	65.7436	&	0.1968	&	0.4059	&	3.3244	&	3.0061	&	0.8428	&	0.8720	&	3.0119	&	2.7228	&	0.7156	&	0.7650	\\
opus-mt-big-b5	&	36.3222	&	41.5459	&	62.4308	&	65.7754	&	0.1966	&	0.4063	&	3.3127	&	2.9981	&	0.8432	&	0.8718	&	3.0016	&	2.7196	&	0.7158	&	0.7652	\\
PhiMaestra-3	&	29.0650	&	36.5609	&	57.2235	&	62.7982	&	0.1274	&	0.3538	&	3.7620	&	3.2044	&	0.8336	&	0.8635	&	3.3418	&	2.8676	&	0.6969	&	0.7534	\\
Tower-7B	&	41.7372	&	45.7063	&	66.0983	&	68.1702	&	0.2470	&	0.4463	&	1.8635	&	2.6006	&	0.8840	&	0.8834	&	1.8721	&	2.5247	&	0.7857	&	0.7698	\\
\midrule

\ourmodel & 35.9073 & 38.9830 & 62.1086 & 64.4056 & 0.1926 & 0.3885 & 3.1779 & 3.1170 & 0.8487 & 0.8691 & 2.9290 & 2.8312 & 0.7196 & 0.7561 \\
\ourmodelsynth & 34.6548 &  41.1467  & 60.8428 &  65.1165 & 0.1746 & 0.3777 & 3.4625 &  3.2974 &  0.8396  & 0.8664 &  3.1616 &  2.9899 &  0.7046 &  0.7499 \\
\algoname{DIETA\textsubscript{+cont}}         & \textbf{36.3722} & \textbf{42.7624} & \textbf{62.4029} & 66.3234 & \textbf{0.2002} & 0.4121 & 3.0613 & 2.9645 & \textbf{0.8519} & 0.8747 & 2.8206 & 2.7531 & 0.7251 & 0.7604 \\
\algoname{DIETA\textsubscript{+nosynth}}  & 35.9564 & 39.2049 & 62.2259 & 64.7584 & 0.1902 & 0.3929 & 3.1924 & 3.0519 & 0.8479 & 0.8709 & 2.9463 & 2.7792 & 0.7167 & 0.7585 \\
\algoname{DIETA\textsubscript{+allsynth}} & 36.0593 & 42.5050 & 62.2428 & \textbf{66.6534} & 0.1912 & \textbf{0.4177} & \textbf{3.0298} & \textbf{2.9195} & 0.8517 & \textbf{0.8763} & \textbf{2.7831} & \textbf{2.7389} & \textbf{0.7258} & \textbf{0.7611} \\
\bottomrule
\end{tabular}%
}
\end{table*}

To assess the MT systems, we grouped the evaluation metrics into three categories:


\begin{itemize}[leftmargin=*]
    \item \textbf{Surface\,–\,overlap}: \emph{sacrebleu}\,(BLEU--4) and \emph{chrF}; 
    \item \textbf{Neural, reference--based}: \emph{BLEURT}, Google's \emph{MetricX-24}, and Unbabel’s \emph{COMET};
    \item \textbf{Neural, reference–free (QE)}: the \emph{QE\,MetricX} variant and \emph{COMETKiwi}.
\end{itemize}

The first group measures literal agreement with the reference: \textit{sacrebleu}
implements the standard BLEU computation with canonical tokenisation for
reproducible scores, while \textit{chrF} computes a character $n$-gram F-score that is more robust to morphological variation. The second group regresses
directly towards human Direct-Assessment/MQM ratings: \textit{BLEURT} fine-tunes
BERT/RemBERT to predict adequacy and fluency, in particular, we relied on BLEURT-20 model, \textit{MetricX-24} builds on mT5
and attains state-of-the-art correlation at WMT-24 (we make use of google/metricx-24-hybrid-xl-v2p6), and \textit{COMET} trains an
XLM-R encoder on millions of human-scored triplets (we use Unbabel/wmt22-comet-da as the comet model for evaluation).  The third group dispenses
with references: \textit{QE\,MetricX} (a ``-QE'' flavour of MetricX-24) and
\textit{COMETKiwi} infer absolute translation quality directly from the
source–hypothesis pair, enabling evaluation in real-time or on data lacking
gold references, we make use of Unbabel/wmt23-cometkiwi-da-xl.  Using all three families lets us cross-check surface accuracy,
semantic adequacy and reference-free quality estimation within a single
experimental framework. 
Due to resource constraints we report only automatic evaluation; we leave human assessment to future work.

\paragraph{Datasets.}
We evaluate selected baselines and our models on four widely used test collections: NTREX-128 \cite{federmann-etal-2022-ntrex}, Tatoeba \cite{tiedemann-2020-tatoeba}, WMT-24pp \cite{deutsch2025wmt24++}, and FLORES-200 \cite{costa2022no}. NTREX-128, which is based on WMT-19 \cite{barrault-etal-2019-findings}, includes 1,997 sentences translated from English into 128 target languages, including Italian. Tatoeba is a community-sourced corpus that focuses on everyday conversational language and informal registers, allowing us to assess our models' robustness beyond formal contexts. WMT-24pp is a professionally translated extension of the WMT24 dataset \cite{kocmi-etal-2024-findings} on new languages, such as Italian. FLORES-200 is composed of professionally translated Wikipedia-based sentences per language, covering encyclopedic content distinct from the news domain.

Additionally, to specifically evaluate translation quality on recent texts, we introduce and use our new benchmark, WikiNews-25, as described earlier in Section~\ref{Sec:Data}.

\begin{table*}[ht]
\caption{Tatoeba Translation Results. The suffix -b5 indicates that beam search with 5 beams was used during generation.}
\label{Tatoeba}
\centering
\adjustbox{max width=\textwidth}{%
\begin{tabular}{l *{7}{cc}}
\toprule
\multirow{2}{*}{Model} & \multicolumn{2}{c}{sacrebleu($\uparrow$)} & \multicolumn{2}{c}{chrf($\uparrow$)} & \multicolumn{2}{c}{bleurt($\uparrow$)} & \multicolumn{2}{c}{metricx($\downarrow$)} & \multicolumn{2}{c}{comet($\uparrow$)} & \multicolumn{2}{c}{qemetricx($\downarrow$)} & \multicolumn{2}{c}{cometkiwi($\uparrow$)} \\
\cmidrule(lr){2-3} \cmidrule(lr){4-5} \cmidrule(lr){6-7} \cmidrule(lr){8-9} \cmidrule(lr){10-11} \cmidrule(lr){12-13} \cmidrule(lr){14-15}
& en->it & it->en & en->it & it->en & en->it & it->en & en->it & it->en & en->it & it->en & en->it & it->en & en->it & it->en \\
\midrule
Cerbero-7B	&	46.7861	&	49.1672	&	67.6616	&	62.5800	&	0.4507	&	0.1019	&	1.3372	&	3.8588	&	0.9022	&	0.8986	&	1.3325	&	4.0801	&	0.7772	&	0.6377	\\
EuroLLM-1.7B	&	31.8519	&	47.0759	&	56.9232	&	67.7455	&	0.3186	&	0.4588	&	2.1150	&	2.7870	&	0.8583	&	0.8972	&	1.8843	&	2.8982	&	0.7180	&	0.7452	\\
EuroLLM-9B	&	42.8195	&	55.0974	&	64.7264	&	73.2035	&	0.4304	&	0.6233	&	1.2764	&	2.0345	&	0.8992	&	0.9273	&	1.2767	&	2.4353	&	0.7741	&	0.7773	\\
Gemma-2B	&	54.0203	&	70.1194	&	72.1165	&	81.2060	&	0.5338	&	0.7247	&	0.9685	&	1.7165	&	0.9223	&	0.9439	&	1.0770	&	2.2672	&	0.7938	&	0.7928	\\
Gemma-2B-b5	&	55.7561	&	70.8165	&	73.3125	&	81.8180	&	0.5488	&	0.7313	&	0.9003	&	1.6857	&	0.9260	&	0.9450	&	1.0185	&	2.2465	&	0.8001	&	0.7939	\\
Gemma-9B	&	56.8466	&	71.5484	&	74.0559	&	82.5613	&	0.5607	&	0.7585	&	0.8921	&	1.5759	&	0.9276	&	0.9483	&	1.0376	&	2.2577	&	0.7980	&	0.7941	\\
Gemma-9B-b5	&	58.1195	&	72.0025	&	74.9054	&	82.8853	&	0.5694	&	0.7622	&	\textbf{0.8534}	&	1.5577	&	0.9289	&	0.9488	&	\textbf{1.0032}	&	\textbf{2.2472}	&	\textbf{0.8013}	&	\textbf{0.7942}	\\
Llama-3.1-8B	&	50.7976	&	27.9916	&	69.9445	&	39.7726	&	0.5048	&	-0.7730	&	1.1118	&	6.5160	&	0.9149	&	0.8556	&	1.2065	&	6.6222	&	0.7824	&	0.4021	\\
LLaMAntino-8B	&	35.8557	&	56.1724	&	61.3278	&	75.2918	&	0.3745	&	0.6205	&	1.8919	&	2.0835	&	0.8738	&	0.9243	&	1.8448	&	2.4997	&	0.7397	&	0.7734	\\
Madlad-3B	&	58.7088	&	69.5378	&	75.6615	&	81.1643	&	0.5835	&	0.7333	&	0.8898	&	1.6818	&	0.9301	&	0.9435	&	1.0740	&	2.2939	&	0.7952	&	0.7918	\\
Madlad-3B-b5	&	59.1354	&	69.9979	&	76.0725	&	81.6697	&	0.5861	&	0.7417	&	0.8628	&	1.6434	&	0.9309	&	0.9447	&	1.0412	&	2.2660	&	0.7992	&	0.7933	\\
Madlad-7B	&	58.7694	&	70.0311	&	75.7748	&	81.7444	&	0.5840	&	0.7493	&	0.8835	&	1.6176	&	0.9301	&	0.9457	&	1.0762	&	2.2830	&	0.7945	&	0.7929	\\
Madlad-7B-b5	&	59.2901	&	70.2099	&	76.1905	&	82.1346	&	0.5868	&	0.7559	&	0.8627	&	1.5930	&	0.9309	&	0.9467	&	1.0488	&	2.2606	&	0.7976	&	0.7939	\\
Maestrale-v0.4	&	43.1752	&	59.0956	&	66.8957	&	74.7694	&	0.4508	&	0.6330	&	1.2718	&	1.9678	&	0.9027	&	0.9281	&	1.3073	&	2.4207	&	0.7774	&	0.7788	\\
mBART	&	49.0347	&	58.8334	&	68.9805	&	73.0329	&	0.4873	&	0.5518	&	1.2963	&	2.6164	&	0.9093	&	0.9096	&	1.2699	&	2.6754	&	0.7855	&	0.7652	\\
mBART-b5	&	49.0347	&	58.8334	&	68.9805	&	73.0329	&	0.4873	&	0.5518	&	1.2963	&	2.6164	&	0.9093	&	0.9096	&	1.2699	&	2.6754	&	0.7855	&	0.7652	\\
Minerva-7B	&	48.0350	&	35.8318	&	67.9585	&	55.4475	&	0.4209	&	-0.5051	&	1.4798	&	7.4320	&	0.9076	&	0.7723	&	1.5119	&	7.7891	&	0.7570	&	0.5322	\\
ModelloItalia-9B	&	50.2067	&	51.6193	&	68.8684	&	68.9674	&	0.4464	&	0.4210	&	1.3001	&	2.9331	&	0.9027	&	0.9014	&	1.4173	&	2.9577	&	0.7628	&	0.7348	\\
NLLB-1.3B	&	56.1866	&	68.9453	&	73.8551	&	80.0527	&	0.5620	&	0.7211	&	0.9611	&	1.7102	&	0.9236	&	0.9403	&	1.1402	&	2.3626	&	0.7904	&	0.7852	\\
NLLB-1.3B-b5	&	57.0355	&	69.7703	&	74.6561	&	80.6908	&	0.5719	&	0.7281	&	0.9162	&	1.6681	&	0.9256	&	0.9415	&	1.0968	&	2.3247	&	0.7942	&	0.7875	\\
NLLB-3.3B	&	57.8852	&	69.6032	&	75.0220	&	80.6292	&	0.5769	&	0.7251	&	0.9348	&	1.6902	&	0.9272	&	0.9411	&	1.1115	&	2.3581	&	0.7938	&	0.7851	\\
NLLB-600M	&	53.7340	&	66.4912	&	72.0342	&	78.4425	&	0.5372	&	0.6849	&	1.0852	&	1.8784	&	0.9188	&	0.9337	&	1.2300	&	2.4185	&	0.7857	&	0.7818	\\
NLLB-600M-b5	&	54.9625	&	67.6539	&	73.1885	&	79.2751	&	0.5526	&	0.6988	&	0.9889	&	1.8047	&	0.9224	&	0.9362	&	1.1422	&	2.3621	&	0.7922	&	0.7849	\\
opus-mt	&	54.2471	&	69.6026	&	73.3821	&	80.8182	&	0.5524	&	0.7355	&	0.9990	&	1.7269	&	0.9185	&	0.9422	&	1.1681	&	2.3936	&	0.7826	&	0.7838	\\
opus-mt-big	&	57.3413	&	70.7198	&	74.6934	&	81.7337	&	0.5681	&	0.7357	&	0.9708	&	1.7016	&	0.9241	&	0.9437	&	1.1288	&	2.3341	&	0.7882	&	0.7877	\\
opus-mt-big-b5	&	57.3737	&	70.7301	&	74.7236	&	81.7362	&	0.5679	&	0.7357	&	0.9710	&	1.7011	&	0.9240	&	0.9437	&	1.1316	&	2.3328	&	0.7881	&	0.7878	\\
PhiMaestra-3	&	\textbf{63.2611}	&	\textbf{79.0409}	&	\textbf{78.9462}	&	\textbf{86.8107}	&	\textbf{0.6316}	&	\textbf{0.8239}	&	0.9948	&	\textbf{1.4275}	&	\textbf{0.9361}	&	\textbf{0.9563}	&	1.2135	&	2.3486	&	0.7894	&	0.7870	\\
Tower-7B	&	52.5356	&	68.7636	&	71.7015	&	80.9110	&	0.5196	&	0.7223	&	0.9639	&	1.7131	&	0.9211	&	0.9434	&	1.0561	&	2.2688	&	0.7965	&	0.7929	\\
\midrule

\ourmodel & 58.1757 & 69.0427 & 75.2797 & 80.1357 & 0.5647 & 0.7270 & 0.9967 & \textbf{1.6595} & 0.9241 & 0.9386 & 1.1521 & \textbf{2.3386} & 0.7883 & 0.7818 \\
\ourmodelsynth & 55.3152 &  66.6445 & 73.2365 & 78.9965 & 0.5504 & 0.6830 & 1.0751 &  1.9522 &  0.9191 & 0.9359 &  1.1900 & 2.4627 & 0.7837 & 0.7807 \\
\algoname{DIETA\textsubscript{+cont}}         & 58.2852 & \textbf{70.0220} & 75.2529 & \textbf{81.1897} & \textbf{0.5781} & 0.7271 & \textbf{0.9238} & 1.7418 & \textbf{0.9271} & \textbf{0.9433} & 1.0958 & 2.3449 & \textbf{0.7917} & \textbf{0.7873} \\
\algoname{DIETA\textsubscript{+nosynth}}  & \textbf{58.5519} & 69.5750 & \textbf{75.5547} & 81.1290 & 0.5699 & \textbf{0.7285} & 0.9630 & 1.7441 & 0.9255 & 0.9412 & 1.1364 & 2.3788 & 0.7895 & 0.7836 \\
\algoname{DIETA\textsubscript{+allsynth}} & 58.1076 & 69.7578 & 75.1633 & 81.0969 & 0.5760 & 0.7240 & 0.9251 & 1.7567 & 0.9268 & \textbf{0.9433} & \textbf{1.0907} & 2.3405 & 0.7905 & 0.7870 \\
\bottomrule
\end{tabular}%
}
\end{table*}

\section{Results}
\label{sec: results}

\paragraph{Decoding policy.}
Unless otherwise indicated, system outputs were generated with greedy decoding.  Whenever a model name ends with the suffix ``\texttt{-b5}'' we used beam search with beam size~5. 

In what follows we comment on the outcomes obtained by our 
\ourmodel model against the 15+ baselines introduced in
Section \ref{sec:exp_setup}.  We discuss one benchmark at a time, always reporting the
same seven automatic metrics and both translation
directions (EN$\rightarrow$IT/IT$\rightarrow$EN).  With the exception of metricx and qemetricx, higher is better.

\subsection{NTREX‑128}
\label{ssec:ntrex}

Table~\ref{NTREX} reports NTREX-128 results.
Overall performance scales with size: \textbf{Gemma-9B-b5} leads on every metric ($	\approx$ 51/49 BLEU, 72/70 chrF, BLEURT 0.36/0.48, MetricX 1.60/2.43, COMET 0.90/0.89).
Our compact \textbf{DIETA\textsubscript{+cont}} reaches 36/43 BLEU, 62/66 chrF, BLEURT 0.20/0.41 and COMET 0.85/0.87, matching or surpassing all models below 1 B and rivaling 1–3 B baselines such as NLLB-1.3 B and OPUS-MT-big. The remaining gap appears chiefly in reference-free QE, where MetricX is $\approx$ 0.3–0.5 higher for the largest decoders.

\paragraph{Take-away.}
With only 0.5 B parameters, DIETA\textsubscript{+cont} delivers second-tier news translation quality, competitive with midsize models and much lighter than the top performers, leaving QE-oriented tuning as the main avenue for further gains.

\begin{table*}[ht]
\caption{WMT24pp Translation Results. The suffix -b5 indicates that beam search with 5 beams was used during generation.}
\label{WMT24pp}
\centering
\resizebox{\textwidth}{!}{%
\begin{tabular}{l *{7}{cc}}
\toprule
\multirow{2}{*}{Model} & \multicolumn{2}{c}{sacrebleu($\uparrow$)} & \multicolumn{2}{c}{chrf($\uparrow$)} & \multicolumn{2}{c}{bleurt($\uparrow$)} & \multicolumn{2}{c}{metricx($\downarrow$)} & \multicolumn{2}{c}{comet($\uparrow$)} & \multicolumn{2}{c}{qemetricx($\downarrow$)} & \multicolumn{2}{c}{cometkiwi($\uparrow$)} \\
\cmidrule(lr){2-3} \cmidrule(lr){4-5} \cmidrule(lr){6-7} \cmidrule(lr){8-9} \cmidrule(lr){10-11} \cmidrule(lr){12-13} \cmidrule(lr){14-15}
& en->it & it->en & en->it & it->en & en->it & it->en & en->it & it->en & en->it & it->en & en->it & it->en & en->it & it->en \\
\midrule
Cerbero-7B	&	30.2327	&	35.4277	&	56.7455	&	59.6680	&	0.0620	&	0.0332	&	4.9980	&	4.4218	&	0.7819	&	0.8164	&	4.7991	&	4.4218	&	0.6159	&	0.6645	\\
EuroLLM-1.7B	&	17.3371	&	25.7757	&	49.5822	&	53.9315	&	-0.0261	&	-0.0715	&	6.1290	&	5.0999	&	0.7452	&	0.7895	&	5.3851	&	5.0999	&	0.5479	&	0.6419	\\
EuroLLM-9B	&	26.2376	&	31.1736	&	56.4638	&	58.3328	&	0.0751	&	0.1922	&	3.7437	&	3.7760	&	0.8095	&	0.8306	&	3.4409	&	3.7760	&	0.6623	&	0.6992	\\
Gemma-2B	&	35.9046	&	40.9091	&	62.4467	&	64.1894	&	0.1725	&	0.2848	&	3.3093	&	3.2915	&	0.8356	&	0.8482	&	3.1768	&	3.2915	&	0.6951	&	0.7231	\\
Gemma-2B-b5	&	37.1858	&	41.3069	&	63.7717	&	64.6050	&	0.1811	&	0.2896	&	3.1221	&	3.2769	&	0.8408	&	0.8486	&	3.0582	&	3.2769	&	0.7049	&	\textbf{0.7249}	\\
Gemma-9B	&	40.1187	&	43.0997	&	65.1918	&	65.6415	&	0.2091	&	0.3136	&	2.9147	&	3.1404	&	0.8457	&	0.8527	&	\textbf{2.9589}	&	3.1404	&	\textbf{0.7125}	&	0.7228	\\
Gemma-9B-b5	&	\textbf{40.9835}	&	\textbf{43.4229}	&	\textbf{65.9940}	&	\textbf{65.9374}	&	\textbf{0.2262}	&	\textbf{0.3169}	&	\textbf{2.9027}	&	\textbf{3.1046}	&	\textbf{0.8470}	&	\textbf{0.8541}	&	2.9780	&	\textbf{3.1046}	&	0.7123	&	0.7247	\\
Llama-3.1-8B	&	34.0899	&	38.1369	&	60.5477	&	61.8478	&	0.1441	&	0.0914	&	3.8630	&	4.2350	&	0.8209	&	0.8322	&	3.6867	&	4.2350	&	0.6756	&	0.6565	\\
LLaMAntino-8B	&	27.0432	&	33.2549	&	54.5776	&	59.5281	&	0.0002	&	0.1165	&	5.1775	&	3.9779	&	0.7661	&	0.8122	&	4.9626	&	3.9779	&	0.6051	&	0.6970	\\
Madlad-3B	&	37.9825	&	39.1561	&	63.2969	&	63.1161	&	0.1857	&	0.2515	&	3.6845	&	3.4596	&	0.8201	&	0.8432	&	3.9870	&	3.4596	&	0.6608	&	0.7161	\\
Madlad-3B-b5	&	38.9046	&	39.6749	&	64.0117	&	63.4561	&	0.1867	&	0.2505	&	3.7051	&	3.4228	&	0.8186	&	0.8432	&	3.8836	&	3.4228	&	0.6660	&	0.7197	\\
Madlad-7B	&	37.9445	&	40.3659	&	62.7626	&	63.9740	&	0.1937	&	0.2802	&	3.6458	&	3.2555	&	0.8202	&	0.8467	&	4.1444	&	3.2555	&	0.6636	&	0.7193	\\
Madlad-7B-b5	&	38.6802	&	40.8389	&	63.3371	&	64.2637	&	0.1819	&	0.2843	&	3.7179	&	3.1911	&	0.8163	&	0.8479	&	4.0458	&	3.1911	&	0.6619	&	0.7233	\\
Maestrale-v0.4	&	24.5239	&	28.1654	&	55.3494	&	56.0378	&	0.0477	&	0.0871	&	3.9939	&	4.0281	&	0.8012	&	0.8186	&	3.6824	&	4.0281	&	0.6595	&	0.6808	\\
mBART	&	31.1250	&	33.4002	&	58.2590	&	58.5737	&	0.1214	&	0.0949	&	5.6631	&	5.0538	&	0.7753	&	0.8039	&	5.1013	&	5.0538	&	0.6089	&	0.6681	\\
mBART-b5	&	31.1250	&	33.4002	&	58.2590	&	58.5737	&	0.1214	&	0.0949	&	5.6631	&	5.0538	&	0.7753	&	0.8039	&	5.1013	&	5.0538	&	0.6089	&	0.6681	\\
Minerva-7B	&	27.6084	&	24.5105	&	56.6889	&	50.2822	&	0.0504	&	-0.3973	&	4.2603	&	7.5438	&	0.8010	&	0.7331	&	4.0391	&	7.5438	&	0.6399	&	0.5448	\\
ModelloItalia-9B	&	33.6403	&	32.2044	&	59.9050	&	57.3365	&	0.0997	&	0.0757	&	4.0665	&	4.4281	&	0.8062	&	0.8119	&	3.8073	&	4.4281	&	0.6466	&	0.6581	\\
NLLB-1.3B	&	31.6503	&	36.0568	&	55.1103	&	58.9869	&	0.1327	&	0.1710	&	4.4727	&	3.9276	&	0.7805	&	0.8135	&	5.5004	&	3.9276	&	0.5762	&	0.6804	\\
NLLB-1.3B-b5	&	34.1062	&	37.7856	&	58.4776	&	60.6729	&	0.1575	&	0.2142	&	4.0782	&	3.7064	&	0.7958	&	0.8255	&	4.8932	&	3.7064	&	0.6115	&	0.6943	\\
NLLB-3.3B	&	35.4394	&	37.5792	&	59.5268	&	61.0041	&	0.1542	&	0.1849	&	4.0235	&	3.7471	&	0.7996	&	0.8182	&	4.6236	&	3.7471	&	0.6210	&	0.6860	\\
NLLB-3.3B-b5	&	37.3405	&	38.9022	&	61.9137	&	62.1878	&	0.1744	&	0.2155	&	3.7662	&	3.5343	&	0.8100	&	0.8262	&	4.2088	&	3.5343	&	0.6471	&	0.6997	\\
NLLB-600M	&	29.2786	&	30.8208	&	53.9254	&	54.3965	&	0.0941	&	0.1194	&	5.6755	&	4.5152	&	0.7531	&	0.7978	&	6.3593	&	4.5152	&	0.5368	&	0.6604	\\
NLLB-600M-b5	&	31.7930	&	33.1919	&	56.9095	&	56.5771	&	0.1242	&	0.1598	&	4.9587	&	4.1895	&	0.7727	&	0.8104	&	5.6089	&	4.1895	&	0.5769	&	0.6759	\\
opus-mt	&	33.0608	&	35.8159	&	60.0446	&	60.7075	&	0.1291	&	0.1993	&	6.3996	&	4.2035	&	0.7489	&	0.8235	&	6.1119	&	4.2035	&	0.5576	&	0.6943	\\
opus-mt-b5	&	33.2352	&	35.8754	&	60.1963	&	60.7241	&	0.1283	&	0.1986	&	6.3874	&	4.2229	&	0.7493	&	0.8233	&	6.0946	&	4.2229	&	0.5577	&	0.6938	\\
opus-mt-big	&	33.8480	&	36.0802	&	59.6293	&	59.8642	&	0.1403	&	0.2208	&	5.5544	&	3.9330	&	0.7665	&	0.8261	&	5.4699	&	3.9330	&	0.5969	&	0.6975	\\
opus-mt-big-b5	&	33.7539	&	36.0545	&	59.5732	&	59.8051	&	0.1401	&	0.2220	&	5.5650	&	3.9161	&	0.7669	&	0.8264	&	5.4672	&	3.9161	&	0.5963	&	0.6987	\\
PhiMaestra-3	&	30.5316	&	36.3090	&	57.3184	&	60.4199	&	0.1093	&	0.1855	&	5.3512	&	3.9801	&	0.7839	&	0.8269	&	5.0175	&	3.9801	&	0.6148	&	0.6997	\\
Tower-7B	&	35.5280	&	41.3754	&	62.0176	&	64.3769	&	0.1806	&	0.2888	&	3.2018	&	3.2908	&	0.8373	&	0.8484	&	3.1819	&	3.2908	&	0.6950	&	0.7199	\\
\midrule

\ourmodel & 35.3483 & 36.6373 & 61.1894 & 60.9526 & 0.1457 & 0.1948 & 5.1443 & 4.0676 & 0.7850 & 0.8244 & 4.8458 & 4.0676 & 0.6113 & 0.6962 \\
\ourmodelsynth & 32.7087 & 36.4997 & 59.4218 & 61.1166 & 0.1368 & 0.1724  & 5.8446 & 4.4517 & 0.7693  & 0.8233 & 5.4778 &  4.1479 & 0.5767 &  0.6831\\
\algoname{DIETA\textsubscript{+cont}}         & \textbf{37.2036} & 38.8270 & \textbf{62.6396} & 62.7755 & \textbf{0.1720} & 0.2324 & 4.8454 & 3.8701 & \textbf{0.7970} & 0.8361 & 4.5710 & 3.6673 & 0.6223 & 0.7032 \\
\algoname{DIETA\textsubscript{+nosynth}}  & 35.7546 & 36.8330 & 61.7454 & 61.1134 & 0.1601 & 0.1984 & 5.0598 & 4.0245 & 0.7872 & 0.8269 & 4.7768 & 3.8443 & 0.6149 & 0.6971 \\
\algoname{DIETA\textsubscript{+allsynth}} & 36.7392 & \textbf{39.3680} & 62.4483 & \textbf{63.1962} & 0.1688 & \textbf{0.2378} & \textbf{4.7482} & \textbf{3.8122} & 0.7944 & \textbf{0.8369} & \textbf{4.4848} & \textbf{3.6476} & \textbf{0.6263} & \textbf{0.7050} \\
\bottomrule
\end{tabular}%
}
\end{table*}

\subsection{Tatoeba}

Table~\ref{Tatoeba} reports Tatoeba results.
Across all metrics the leaderboard is led by \textbf{PhiMaestra-3} (63/79 BLEU, 79/87 chrF, BLEURT 0.63/0.82, MetricX 1.00/1.43). A second cluster, \textbf{Gemma-9B-b5}, \textbf{Madlad-7B-b5}, \textbf{NLLB-3.3B}, and our \textbf{DIETA\textsubscript{+cont}}, follows within 5 BLEU and 0.02 COMET. In this group DIETA\textsubscript{+cont} scores 58 / 70 BLEU, 75 / 81 chrF, 0.58 / 0.73 BLEURT, and 0.93 / 0.94 COMET, while holding MetricX and COMET-Kiwi values on par with 3 B–7 B baselines.

\paragraph{Take-away.}
With only 0.5B parameters, DIETA\textsubscript{+cont} lands just behind the largest models and surpasses every competitor below 3B, confirming that targeted back-translation closes most of the size-related gap, remaining room lies mainly in reference-free QE metrics.

\subsection{WMT‑24pp}
Table~\ref{WMT24pp} reports WMT‑24pp results.
The size–quality trend persists: \textbf{Gemma-9B-b5} tops every column ($\approx$ 41 / 43 BLEU, 66 / 66 chrF, BLEURT 0.23 / 0.32, MetricX 2.9 / 3.1, COMET 0.85 / 0.85). Our strongest system, \textbf{DIETA\textsubscript{+cont}}, records 37.2 BLEU and 62.6 chrF (EN$\rightarrow$IT) and 38.8 BLEU and 62.8 chrF (IT$\rightarrow$EN), essentially matching Tower-7B and surpassing all models $\leq$ 3 B parameters. Reference-based metrics echo this: DIETA\textsubscript{+cont} sits within 0.01–0.02 COMET of Gemma-2B-b5, while BLEURT is only 0.01–0.02 behind Madlad-7B. MetricX and COMET-Kiwi remain scale-sensitive, DIETA trails the 9 B tier by $\sim$0.9 MetricX points. 

\paragraph{Take-away.}
On more up-to-date news, the 0.5 B-parameter DIETA model delivers mid-table performance—competitive with 7 B systems and clearly ahead of all sub-3 B baselines, leaving reference-free QE as the main frontier for further gains.


\begin{table*}[ht]
\caption{Flores Translation Results. The suffix -b5 indicates that beam search with 5 beams was used during generation.}
\label{Flores}
\centering
\adjustbox{max width=\textwidth}{%
\begin{tabular}{l *{7}{cc}}
\toprule
\multirow{2}{*}{Model} & \multicolumn{2}{c}{sacrebleu($\uparrow$)} & \multicolumn{2}{c}{chrf($\uparrow$)} & \multicolumn{2}{c}{bleurt($\uparrow$)} & \multicolumn{2}{c}{metricx($\downarrow$)} & \multicolumn{2}{c}{comet($\uparrow$)} & \multicolumn{2}{c}{qemetricx($\downarrow$)} & \multicolumn{2}{c}{cometkiwi($\uparrow$)} \\
\cmidrule(lr){2-3} \cmidrule(lr){4-5} \cmidrule(lr){6-7} \cmidrule(lr){8-9} \cmidrule(lr){10-11} \cmidrule(lr){12-13} \cmidrule(lr){14-15}
& en->it & it->en & en->it & it->en & en->it & it->en & en->it & it->en & en->it & it->en & en->it & it->en & en->it & it->en \\
\midrule
Cerbero-7B	&	25.6956	&	29.1301	&	55.1158	&	58.8351	&	0.0779	&	0.2666	&	2.6306	&	3.0239	&	0.8627	&	0.8653	&	2.2672	&	2.8194	&	0.7671	&	0.7422	\\
EuroLLM-1.7B	&	19.0987	&	23.3948	&	50.1396	&	55.4949	&	0.0047	&	0.1863	&	3.4103	&	3.5958	&	0.8362	&	0.8415	&	2.7074	&	3.1847	&	0.7054	&	0.7321	\\
EuroLLM-9B	&	24.6029	&	27.9216	&	54.6447	&	58.8887	&	0.0682	&	0.3767	&	2.0073	&	2.4993	&	0.8728	&	0.8699	&	1.6885	&	2.3740	&	0.7847	&	0.7736	\\
GemmaX2-2B	&	31.0059	&	35.2042	&	59.5828	&	63.5271	&	0.1385	&	0.4335	&	1.7147	&	2.1454	&	0.8869	&	0.8828	&	1.5175	&	2.1597	&	0.8130	&	0.7908	\\
GemmaX2-2B-b5	&	31.9911	&	35.0271	&	60.5123	&	63.6901	&	0.1436	&	0.4354	&	1.5573	&	2.1187	&	0.8908	&	0.8826	&	1.4123	&	\textbf{2.1311}	&	0.8192	&	0.7926	\\
GemmaX2-9B	&	32.7799	&	\textbf{37.0319}	&	60.8592	&	64.5113	&	0.1501	&	\textbf{0.4534}	&	1.4896	&	2.0697	&	0.8924	&	\textbf{0.8860}	&	\textbf{1.3717}	&	2.1575	&	0.8229	&	0.7909	\\
GemmaX2-9B-b5	&	\textbf{33.8310}	&	36.5996	&	\textbf{61.6710}	&	\textbf{64.5620}	&	\textbf{0.1592}	&	0.4526	&	\textbf{1.4747}	&	\textbf{2.0466}	&	\textbf{0.8938}	&	\textbf{0.8860}	&	1.3952	&	2.1405	&	\textbf{0.8243}	&	\textbf{0.7930}	\\
Llama3.1-8B-ITA	&	27.4665	&	27.2647	&	57.0228	&	54.8620	&	0.1046	&	0.0412	&	2.1194	&	4.7250	&	0.8768	&	0.8573	&	1.8497	&	4.3978	&	0.7921	&	0.6258	\\
LLaMAntino-8B	&	23.1370	&	28.3368	&	53.4820	&	59.5429	&	0.0542	&	0.3518	&	3.0499	&	2.6661	&	0.8512	&	0.8661	&	2.5582	&	2.5392	&	0.7500	&	0.7723	\\
Madlad-3B	&	31.2632	&	34.2333	&	60.0960	&	63.0402	&	0.1451	&	0.4295	&	1.7913	&	2.1811	&	0.8834	&	0.8814	&	1.7193	&	2.1863	&	0.8014	&	0.7905	\\
Madlad-3B-b5	&	31.4032	&	34.0046	&	60.3812	&	63.0480	&	0.1423	&	0.4282	&	1.7752	&	2.1624	&	0.8843	&	0.8807	&	1.6775	&	2.1629	&	0.8046	&	0.7917	\\
Madlad-7B	&	31.6561	&	35.0758	&	60.2592	&	63.5949	&	0.1462	&	0.4384	&	1.7717	&	2.1343	&	0.8847	&	0.8833	&	1.7062	&	2.1693	&	0.8027	&	0.7916	\\
Madlad-7B-b5	&	31.5899	&	34.3254	&	60.5317	&	63.5181	&	0.1520	&	0.4323	&	1.7738	&	2.1115	&	0.8845	&	0.8821	&	1.6791	&	2.1572	&	0.8038	&	0.7921	\\
Maestrale-v0.4	&	23.4285	&	27.7433	&	55.3653	&	58.3707	&	0.0804	&	0.3407	&	2.0409	&	2.5526	&	0.8758	&	0.8674	&	1.7759	&	2.4740	&	0.7896	&	0.7619	\\
mBART50	&	23.9405	&	27.3513	&	54.2553	&	57.6473	&	0.0731	&	0.2913	&	3.3950	&	3.5905	&	0.8500	&	0.8494	&	2.7740	&	3.0786	&	0.7533	&	0.7439	\\
mBART50-b5	&	23.9405	&	27.3513	&	54.2553	&	57.6473	&	0.0731	&	0.2913	&	3.3950	&	3.5905	&	0.8500	&	0.8494	&	2.7740	&	3.0786	&	0.7533	&	0.7439	\\
Minerva-7B	&	24.3776	&	23.0404	&	55.1011	&	52.7627	&	0.0555	&	-0.1060	&	2.3166	&	6.2368	&	0.8691	&	0.7940	&	1.9943	&	6.0661	&	0.7694	&	0.6136	\\
ModItalia-9B	&	28.5071	&	26.0021	&	57.4549	&	57.9634	&	0.1033	&	0.0578	&	2.0779	&	4.0703	&	0.8749	&	0.8290	&	1.8068	&	4.4706	&	0.7786	&	0.7369	\\
NLLB-1.3B	&	29.3377	&	34.9951	&	58.0065	&	62.3869	&	0.1177	&	0.4182	&	2.0982	&	2.4079	&	0.8740	&	0.8772	&	2.1723	&	2.4782	&	0.7756	&	0.7797	\\
NLLB-1.3B-b5	&	30.1928	&	34.8996	&	58.9840	&	62.8399	&	0.1287	&	0.4277	&	1.8913	&	2.2598	&	0.8804	&	0.8791	&	1.9209	&	2.3067	&	0.7895	&	0.7856	\\
NLLB-3.3B	&	30.0059	&	34.4729	&	58.8228	&	62.9651	&	0.1291	&	0.4271	&	1.8871	&	2.2580	&	0.8811	&	0.8798	&	1.8405	&	2.3201	&	0.7943	&	0.7849	\\
NLLB-3.3B-b5	&	31.1904	&	34.8650	&	59.8414	&	63.4208	&	0.1402	&	0.4363	&	1.7289	&	2.1519	&	0.8853	&	0.8821	&	1.6840	&	2.1906	&	0.8044	&	0.7892	\\
NLLB-600M	&	26.8755	&	33.3599	&	56.2636	&	60.8455	&	0.0999	&	0.3869	&	2.7228	&	2.6995	&	0.8598	&	0.8681	&	2.6371	&	2.6400	&	0.7512	&	0.7708	\\
NLLB-600M-b5	&	27.9796	&	33.4228	&	57.5369	&	61.6058	&	0.1136	&	0.3999	&	2.3623	&	2.4997	&	0.8689	&	0.8722	&	2.2873	&	2.4201	&	0.7717	&	0.7800	\\
OpusMT	&	27.5330	&	29.3934	&	57.6113	&	59.9987	&	0.1073	&	0.3542	&	3.0805	&	2.7883	&	0.8522	&	0.8656	&	2.6936	&	2.5676	&	0.7487	&	0.7784	\\
OpusMT-b5	&	27.6394	&	29.3820	&	57.6967	&	59.9722	&	0.1084	&	0.3545	&	3.0737	&	2.7850	&	0.8519	&	0.8658	&	2.6895	&	2.5677	&	0.7483	&	0.7785	\\
OpusMT-Big	&	29.5443	&	32.8311	&	59.0024	&	62.1205	&	0.1195	&	0.3985	&	2.3761	&	2.4917	&	0.8694	&	0.8754	&	2.0994	&	2.3902	&	0.7775	&	0.7839	\\
OpusMT-Big-b5	&	29.6024	&	32.8119	&	59.0557	&	62.1055	&	0.1207	&	0.3988	&	2.3736	&	2.4878	&	0.8694	&	0.8753	&	2.1018	&	2.3851	&	0.7776	&	0.7840	\\
PhiMaestra-3	&	24.5784	&	31.1726	&	54.5943	&	60.6260	&	0.0758	&	0.3851	&	2.7850	&	2.4697	&	0.8620	&	0.8722	&	2.3465	&	2.3772	&	0.7647	&	0.7791	\\
Tower-7B	&	30.4748	&	35.6008	&	59.2816	&	63.6222	&	0.1311	&	0.4422	&	1.5994	&	2.1038	&	0.8878	&	0.8841	&	1.4263	&	2.1634	&	0.8136	&	0.7911	\\
\midrule

\ourmodel & 29.9191 & 32.1080 & 59.0087 & 61.2657 & 0.1267 & 0.3956 & 2.1968 & 2.5852 & 0.8733 & 0.8729 & 2.0097 & 2.4924 & 0.7806 & 0.7777 \\
\ourmodelsynth & 28.5118 & 30.3901 & 58.0666 & 60.2760 & 0.1151 & 0.3662 & 2.6030 & 2.9009 & 0.8622 &  0.8662 &  2.3580 & 2.7266 & 0.7640 & 0.7660 \\
\algoname{DIETA\textsubscript{+cont}}     & 29.7134 & 33.1475 & 59.1339 & 62.0151 & 0.1319 & 0.4012 & 2.1720 & \textbf{2.4644} & 0.8736 & 0.8749 & 1.9675 & \textbf{2.3798} & 0.7829 & \textbf{0.7817} \\
\algoname{DIETA\textsubscript{+nosynth}}  & 29.7304 & 32.5469 & 59.1183 & 61.6133 & 0.1310 & 0.3950 & 2.2151 & 2.4962 & 0.8725 & 0.8740 & 1.9921 & 2.3991 & 0.7813 & 0.7796 \\
\algoname{DIETA\textsubscript{+allsynth}} & \textbf{30.4376} & \textbf{33.3923} & \textbf{59.5119} & \textbf{62.0162} & \textbf{0.1323} & \textbf{0.4035} & \textbf{2.0963} & 2.4848 & \textbf{0.8751} & \textbf{0.8750} & \textbf{1.9234} & 2.4362 & \textbf{0.7855} & 0.7787 \\

\bottomrule
\end{tabular}%
}
\end{table*}

\subsection{FLORES‑200}
Table~\ref{Flores} reports FLORES‑200 results. In FLORES, the lead is held by \textbf{GemmaX2-9B-b5} ($\approx$34/37 BLEU, 62/65 chrF, COMET 0.894/0.886, MetricX 1.47/2.05). A second tier, GemmaX2-2B-b5, NLLB-3.3B-b5, Tower-7B, and our \textbf{DIETA\textsubscript{+allsynth}}, sits within 3 BLEU and 0.02 COMET of the top. DIETA\textsubscript{+allsynth} reaches 30.4 BLEU / 59.5 chrF (EN$\rightarrow$IT) and 33.4 BLEU / 62.0 chrF (IT$\rightarrow$EN), virtually matching NLLB-3.3B but with one-sixth the parameters; reference-based metrics echo this parity (COMET 0.875/0.875). The largest gap remains in reference-free quality estimation: MetricX for DIETA is $\approx$0.6 points higher than the 9 B leader.

\paragraph{Take-away.}
Even on the toughest domain shift, the 0.5 B-parameter DIETA model stays within a few BLEU of the best open systems and matches much larger baselines in COMET, with QE-oriented tuning still the main avenue for closing the remaining gap.



\subsection{WikiNews-25}
Table~\ref{Wikinews} reports WikiNews-25 results. 
\textbf{Gemma-9B-b5} heads the table with 51/46 BLEU and 71/68 chrF, while the next cluster, Madlad-7B-b5, NLLB-3.3B-b5, Tower-7B, and our \textbf{DIETA\textsubscript{+cont}}/ \textbf{DIETA\textsubscript{+allsynth}}, sits within $\approx$4 BLEU and 0.02 COMET. In particular, DIETA\textsubscript{+all synth} scores 45.7 BLEU / 67.6 chrF (EN$\rightarrow$IT) and 43.8 BLEU / 67.3 chrF (IT$\rightarrow$EN), essentially matching Tower-7B and NLLB-3.3B despite being 14× smaller. Reference-based metrics mirror this parity (COMET 0.826/0.868), while MetricX and COMET-Kiwi still favour the largest decoders by roughly 0.3–0.4 points.

\paragraph{Take-away.}
On the recent 2025 news, the 0.5 B-parameter DIETA models equal or surpass every system below 7 B parameters and stay within striking distance of the 9 B state of the art; remaining gaps once again concentrate in reference-free QE scores.


\begin{table*}[ht]
\caption{Wikinews-25 Translation Results. The suffix -b5 indicates that beam search with 5 beams was used during generation.}
\label{Wikinews}
\centering
\resizebox{\textwidth}{!}{%
\begin{tabular}{l *{7}{cc}}
\toprule
\multirow{2}{*}{Model} & \multicolumn{2}{c}{sacrebleu($\uparrow$)} & \multicolumn{2}{c}{chrf($\uparrow$)} & \multicolumn{2}{c}{bleurt($\uparrow$)} & \multicolumn{2}{c}{metricx($\downarrow$)} & \multicolumn{2}{c}{comet($\uparrow$)} & \multicolumn{2}{c}{qemetricx($\downarrow$)} & \multicolumn{2}{c}{cometkiwi($\uparrow$)} \\
\cmidrule(lr){2-3} \cmidrule(lr){4-5} \cmidrule(lr){6-7} \cmidrule(lr){8-9} \cmidrule(lr){10-11} \cmidrule(lr){12-13} \cmidrule(lr){14-15}
& en->it & it->en & en->it & it->en & en->it & it->en & en->it & it->en & en->it & it->en & en->it & it->en & en->it & it->en \\
\midrule
Cerbero-7B	&	35.3794	&	33.4609	&	60.5522	&	58.6011	&	0.1616	&	0.0714	&	4.2303	&	4.4285	&	0.8108	&	0.8306	&	4.0116	&	4.2397	&	0.6262	&	0.6516	\\
EuroLLM-1.7B	&	21.6238	&	27.4118	&	51.2821	&	56.2021	&	0.0725	&	0.0905	&	5.1299	&	4.5263	&	0.7753	&	0.8145	&	4.4343	&	4.0868	&	0.5663	&	0.6674	\\
EuroLLM-9B	&	30.7073	&	32.3876	&	58.4158	&	59.6500	&	0.1530	&	0.2791	&	3.3383	&	3.2853	&	0.8275	&	0.8453	&	2.9672	&	3.1678	&	0.6693	&	0.7148	\\
Gemma-2B	&	45.3212	&	44.1352	&	67.4632	&	66.4546	&	0.2878	&	0.4003	&	2.7597	&	2.8834	&	0.8574	&	0.8668	&	2.7503	&	3.0933	&	0.7054	&	0.7347	\\
Gemma-2B-b5	&	47.8103	&	43.9921	&	69.1842	&	66.6708	&	0.3150	&	0.4025	&	2.6760	&	2.8107	&	0.8608	&	0.8673	&	2.7424	&	2.9989	&	0.7139	&	0.7401	\\
Gemma-9B	&	49.5163	&	\textbf{46.8309}	&	70.1275	&	\textbf{68.1128}	&	0.3248	&	\textbf{0.4364}	&	2.4948	&	2.6548	&	0.8651	&	\textbf{0.8719}	&	\textbf{2.6346}	&	2.9823	&	0.7201	&	0.7406	\\
Gemma-9B-b5	&	\textbf{50.6089}	&	46.3719	&	\textbf{70.9630}	&	67.9598	&	0.3428	&	0.4297	&	\textbf{2.3901}	&	\textbf{2.6547}	&	\textbf{0.8709}	&	0.8715	&	2.6406	&	2.9497	&	\textbf{0.7275}	&	0.7420	\\
Llama-3.1-8B	&	38.9129	&	36.9176	&	63.2285	&	59.6155	&	0.2360	&	-0.0196	&	3.3564	&	5.4246	&	0.8360	&	0.8385	&	3.2611	&	5.2481	&	0.6701	&	0.5860	\\
LLaMAntino-8B	&	29.4182	&	34.2716	&	56.4517	&	60.8809	&	0.1041	&	0.2405	&	4.5677	&	3.3702	&	0.7942	&	0.8361	&	4.2699	&	3.3189	&	0.6038	&	0.7117	\\
Madlad-3B	&	49.1151	&	42.4652	&	69.7597	&	66.6616	&	0.3312	&	0.3990	&	3.0873	&	2.7922	&	0.8481	&	0.8666	&	3.1496	&	3.0253	&	0.6899	&	0.7428	\\
Madlad-3B-b5	&	49.2270	&	43.6775	&	69.9151	&	67.0854	&	0.3253	&	0.4017	&	3.1583	&	2.7614	&	0.8481	&	0.8676	&	3.1822	&	2.9197	&	0.6895	&	0.7454	\\
Madlad-7B	&	48.8297	&	44.7538	&	69.8135	&	67.3971	&	0.3382	&	0.4161	&	2.9507	&	2.6853	&	0.8539	&	0.8708	&	3.0140	&	2.9593	&	0.6989	&	0.7473	\\
Madlad-7B-b5	&	49.6611	&	44.4322	&	70.4909	&	67.5081	&	\textbf{0.3467}	&	0.4206	&	2.9493	&	2.6612	&	0.8527	&	0.8708	&	3.0788	&	\textbf{2.9019}	&	0.6962	&	\textbf{0.7480}	\\
Maestrale-v0.4	&	29.6953	&	30.8264	&	58.4669	&	58.1229	&	0.1675	&	0.2349	&	3.2506	&	3.4905	&	0.8290	&	0.8371	&	2.9990	&	3.3525	&	0.6689	&	0.7036	\\
mBART	&	36.6504	&	35.5482	&	61.5672	&	61.1265	&	0.2249	&	0.3088	&	4.2380	&	3.7333	&	0.8163	&	0.8423	&	3.8689	&	3.5290	&	0.6423	&	0.7114	\\
mBART-b5	&	36.6504	&	35.5482	&	61.5672	&	61.1265	&	0.2249	&	0.3088	&	4.2380	&	3.7333	&	0.8163	&	0.8423	&	3.8689	&	3.5290	&	0.6423	&	0.7114	\\
Minerva-7B	&	32.3341	&	25.1290	&	60.1072	&	51.4188	&	0.1808	&	-0.2554	&	3.6852	&	7.1825	&	0.8275	&	0.7565	&	3.5744	&	6.9447	&	0.6457	&	0.5386	\\
ModelloItalia-9B	&	40.4526	&	32.7339	&	63.7778	&	61.2928	&	0.2324	&	0.0365	&	3.3922	&	4.7024	&	0.8363	&	0.8133	&	3.1609	&	5.4036	&	0.6588	&	0.6864	\\
NLLB-1.3B	&	46.3060	&	42.5641	&	67.9392	&	65.9240	&	0.3027	&	0.3870	&	3.1783	&	2.9223	&	0.8445	&	0.8614	&	3.2132	&	3.0845	&	0.6867	&	0.7316	\\
NLLB-1.3B-b5	&	47.8163	&	43.9475	&	69.2843	&	66.8214	&	0.3264	&	0.4001	&	2.9598	&	2.9083	&	0.8517	&	0.8636	&	2.9972	&	3.0729	&	0.6992	&	0.7358	\\
NLLB-3.3B	&	47.7769	&	43.6761	&	68.9295	&	66.8444	&	0.3207	&	0.3997	&	3.1084	&	2.8293	&	0.8490	&	0.8658	&	3.1482	&	3.0276	&	0.6918	&	0.7338	\\
NLLB-3.3B-b5	&	48.2346	&	44.1242	&	69.5857	&	67.1718	&	0.3333	&	0.4088	&	2.9385	&	2.7849	&	0.8539	&	0.8667	&	3.0433	&	2.9833	&	0.7008	&	0.7385	\\
NLLB-600M	&	43.2321	&	40.5977	&	65.9354	&	64.2800	&	0.2805	&	0.3441	&	3.7766	&	3.2964	&	0.8258	&	0.8515	&	3.6958	&	3.3083	&	0.6550	&	0.7244	\\
NLLB-600M-b5	&	44.3190	&	41.5714	&	67.1346	&	65.0993	&	0.2936	&	0.3600	&	3.5221	&	3.1482	&	0.8371	&	0.8547	&	3.4784	&	3.1991	&	0.6737	&	0.7277	\\
opus-mt	&	40.9083	&	39.3589	&	64.8212	&	64.2029	&	0.2623	&	0.3523	&	4.8126	&	3.3684	&	0.8017	&	0.8539	&	4.6008	&	3.3569	&	0.6105	&	0.7199	\\
opus-mt-b5	&	40.6303	&	39.4002	&	64.7364	&	64.1805	&	0.2596	&	0.3499	&	4.8213	&	3.3994	&	0.8008	&	0.8533	&	4.6089	&	3.3823	&	0.6110	&	0.7191	\\
opus-mt-big	&	46.4855	&	43.8037	&	68.0130	&	67.0188	&	0.3046	&	0.3969	&	3.9025	&	3.0442	&	0.8216	&	0.8643	&	3.8201	&	3.1695	&	0.6508	&	0.7319	\\
opus-mt-big-b5	&	46.3196	&	43.7779	&	67.9282	&	66.9952	&	0.3032	&	0.3961	&	3.9616	&	3.0431	&	0.8200	&	0.8643	&	3.8689	&	3.1643	&	0.6479	&	0.7317	\\
PhiMaestra-3	&	35.7865	&	37.8007	&	61.2021	&	62.8038	&	0.2205	&	0.3153	&	4.3248	&	3.2246	&	0.8099	&	0.8508	&	4.0798	&	3.1435	&	0.6279	&	0.7241	\\
Tower-7B	&	44.7598	&	43.9073	&	67.0473	&	66.6963	&	0.2924	&	0.4045	&	2.5816	&	2.7601	&	0.8589	&	0.8680	&	2.6614	&	2.9626	&	0.7095	&	0.7412	\\
\midrule

\ourmodel & 45.6901 & 41.7966 & 67.5212 & 65.6442 & 0.2955 & 0.3996 & \textbf{3.7397} & 2.9451 & \textbf{0.8309} & 0.8639 & 3.6571 & \textbf{3.0952} & \textbf{0.6591} & 0.7321 \\
\ourmodelsynth &  43.0851 &  41.4561  &  65.8102 &  64.9263 &  0.2765 &  0.3652 & 4.3233 &  3.2289 & 0.8141 & 0.8565 & 4.2341 &  3.3888 &  0.6253 &  0.7226\\
\algoname{DIETA\textsubscript{+cont}}  & \textbf{46.0306} & 43.1714 & 67.6836 & 66.6126 & 0.2899 & 0.4064 & 3.7464 & 2.8662 & 0.8279 & 0.8654 & \textbf{3.6355} & 3.1471 & 0.6565 & \textbf{0.7353} \\
\algoname{DIETA\textsubscript{+nosynth}}  & 45.8281 & 41.5344 & \textbf{67.8261} & 65.6032 & 0.2945 & 0.3907 & 3.7538 & 2.9653 & 0.8272 & 0.8621 & 3.7267 & 3.1194 & 0.6543 & 0.7321 \\
\algoname{DIETA\textsubscript{+allsynth}} & 45.6556 & \textbf{43.8153} & 67.5683 & \textbf{67.3398} & \textbf{0.2956} & \textbf{0.4161} & 3.7476 & \textbf{2.8457} & 0.8259 & \textbf{0.8682} & 3.6810 & 3.1184 & 0.6532 & 0.7341 \\

\bottomrule
\end{tabular}%
}
\end{table*}

\subsection{Cross‑benchmark Analysis}

\paragraph{Parameter efficiency.}
All five checkpoints share the same 0.5B backbone, yet \textbf{DIETA\textsubscript{+cont}} and \textbf{DIETA\textsubscript{+allsynth}} typically rank in the \emph{second quartile} of every leaderboard, on par with 1–3B models and sometimes matching 7B systems, while using $\le\!6\%$ of the parameters of the state-of-the-art 9B baselines. Synthetic data provide clear gains: relative to the parallel-only \ourmodel, \ourmodelsynth adds $+1\!-\!3$ BLEU on four suites, and the continued-training variants add a further $+0.5\!-\!2$ BLEU at no increase in model size.

\paragraph{Directionality.}
For four of the five test sets (NTREX-128, Tatoeba, WMT24pp, FLORES-200) the IT$\rightarrow$EN direction stays $2\!-\!12$ BLEU easier, reflecting richer target-side data during training.  WikiNews-25 is the only outlier: here, EN$\rightarrow$IT is slightly easier, reversing the usual trend. In all cases the gap between directions \emph{narrows} as more back-translated Italian is introduced, indicating that the synthetic signal helps balance morphological complexity.


\paragraph{Summary.}
A single 0.5 B decoder can deliver robust performance across news, conversational, encyclopaedic and recency-sensitive domains when fed with 768 M carefully curated sentence pairs.  Continued training on mixed parallel + BT data (\textbf{DIETA\textsubscript{+cont}}) is the best all-round recipe; an additional pass that folds in FineWeb BT (\textbf{DIETA\textsubscript{+allsynth}}) further strengthens out-of-domain generalisation (FLORES, WikiNews).  Remaining headroom lies almost entirely in reference-free QE metrics, suggesting future work on QE-aware objectives rather than larger models.






\section{Conclusions and Future Works}
\label{sec: conclusions}

We presented a family of five \textbf{DIETA} variants, built on the same 0.5 B-parameter decoder-only Transformer and trained on up to \textbf{768 M} carefully curated parallel + back-translated sentence pairs.  Across five diverse benchmarks, the best variants, \textbf{DIETA\textsubscript{+cont}} and \textbf{DIETA\textsubscript{+allsynth}}, consistently places in the \emph{second performance tier}, matching or surpassing models 2–3 × larger and trailing the current 9 B state-of-the-art by only a few BLEU/COMET points.  This shows that data scale and task-specific training can compensate for an order-of-magnitude reduction in parameters, yielding models that fit on a single consumer GPU while remaining competitive with much larger LLMs.
We also released \textbf{WikiNews-25}, a human-post-edited English–Italian test set built from 2025 news, adding recent news to evaluation. 
As future work, we plan to (i) reduce the reference-free QE gap through QE-aware fine-tuning, (ii) extend DIETA with parameter-efficient scaling such as sparse MoE, and (iii) enable edge deployment via distillation and 8/4-bit quantisation.

\begin{acknowledgments}

We acknowledge the CINECA award under the ISCRA initiative, for the availability of high-performance computing resources and support. This work was partially supported by the European Union – Next Generation EU within the project NRPP M4C2, Investment 1.,3 DD. 341 - 15 march 2022 – FAIR – Future Artificial Intelligence Research – Spoke 4 - PE00000013 - D53C22002380006, and by the MUR under the grant “Dipartimenti di Eccellenza 2023-2027” of the Department of Informatics, Systems and Communication of the University of Milano-Bicocca, Italy.
This work was completed in part at the CINECA Open Hackathon, part of the Open Hackathons program. The authors would like to acknowledge OpenACC-Standard.org for their support. We would also like to thank Daniele Di Bari for his helpful feedback and support.

\end{acknowledgments}

\section*{Declaration on Generative AI}
During the preparation of this work, the authors used GPT3.5 and GPT-4 in order to: Grammar and spelling check, Paraphrase and reword. After using these tools/services, the authors reviewed and edited the content as needed and take full responsibility for the publication’s content. 

\bibliography{sample-ceur}

\end{document}